\documentclass[12pt]{article}
\usepackage[margin=1in]{geometry}
\usepackage[T1]{fontenc}
\usepackage{mathptmx}
\usepackage{amsmath}
\usepackage{amssymb}
\usepackage{graphicx}
\usepackage{natbib}
\usepackage{hyperref}
\hypersetup{hidelinks}
\usepackage{booktabs}
\usepackage{array}
\usepackage{multirow}
\usepackage{microtype}
\usepackage{setspace}
\title{Change My View? The Dynamics of Persuasion and Polarization in Online Discourse}
\author{ David Freeborn, Malihe Alikhani, Anthony Sicilia}
\date{}

\begin{document}

\maketitle

\begin{abstract}
Philosophical accounts of persuasion often assume that shared evidence and rational argumentation should lead to a convergence of views between peers, yet everyday discourse often suggests otherwise. In this study, we use large language models to analyze a corpus of debates on Reddit's r/ChangeMyView, where belief revision is publicly signaled. Large language models were asked, halfway through each discussion, to forecast whether such an acknowledgement would arise; their probabilistic estimates serve as a conversational baseline. Each reply was then coded, through a hybrid machine-assisted procedure, for ten familiar rhetorical strategies---concession, empathy, logical challenge, credibility appeals, and so forth. Adding these strategic features markedly improves predictive power and yields a consistent pattern: moves that express concession or empathetic alignment substantially increase the prospect of belief change, whereas frontal refutation, credibility attacks, and topic deflection diminish it. The findings indicate that effective public reasoning depends as much on relational framing as on evidential content, and they invite a refinement of normative accounts of rational dialogue.
\end{abstract}

\section{Introduction: The Puzzle of Persuasion and Polarization}

On 27 September 2018 the United States paused to watch Christine Blasey Ford and Brett Kavanaugh testify before the Senate Judiciary Committee. Before the hearing an NPR/PBS NewsHour/Marist poll \citep{marist2018sept} found the public largely undecided: 42\% were unsure whom to believe, while belief in Ford (32\%) and in Kavanaugh (26\%) was roughly balanced. However, just days later after watching the same testimony, the undecided share had collapsed to 22\% and opinions had pulled apart: belief in Ford jumped to 45\%, belief in Kavanaugh nudged up to 33\%. Further polling suggested only 5\% of respondents now judged both witnesses credible; instead, 73\% of Democrats believed Ford while 74\% of Republicans believed Kavanaugh \citep{marist2018oct}. Thus the very moment designed to elicit a shared judgment instead pushed the poles of opinion further apart.

The Kavanaugh case exemplifies a puzzle for philosophers. The same body of evidence---two sworn testimonies, a single televised exchange---generated sharply divergent posterior beliefs, challenging the expectation that rational agents exposed to common data should converge. The episode illustrates the double-edged nature of public reasoning: whereas deliberation is often lauded as a pathway to collective truth, here it seemed to entrench factional narratives, suggesting that sincere, information-rich discourse can rationally widen disagreement when background credences are entangled.

Philosophers often expect rational inquiry to be truth-conducive, yet in cases like the Kavanaugh testimonials, it sometimes only exacerbates disagreement. This raises what we shall call the ``puzzle of persuasion'': why does rational persuasion sometimes succeed whilst at other times it can fail so spectacularly? Classical Bayesian epistemology appears to promise an answer: agents with absolutely continuous priors who update on the same data must eventually merge their opinions \citep{blackwell1962}. Yet the Ford--Kavanaugh case, and countless smaller exchanges, belie that promise. Indeed, recent formal work \citep{freeborn2023, freeborn2024} proves the fragility of merger: when agents hold multiple, probabilistically entangled beliefs, Bayesian updating can rationally generate persistent polarisation and even predictable factionalisation.

This paper tackles the puzzle of persuasion with an empirical lens, using a novel philosophical methodology. We analyse more than 3,000 discussions from Reddit's Change My View forum, where original posters publicly signal persuasion by awarding a delta badge. Our approach builds on methods developed by \citet{sicilia2024}, who show that pre-trained language models can forecast conversational outcomes and, when calibrated for uncertainty, markedly reduce bias in prediction.

As such we set out to answer three related questions. First, can large language models help us to forecast persuasion? We prompt Llama-3 variants to assign probabilities that an unfinished Change My View thread will end in a badge. Second, which argumentative strategies actually work? We use two complementary taxonomies, one distilled from the social-influence literature, the other induced by chain-of-thought prompting: we label every reply for strategic content (concession, logical appeal, shared-identity cues, pathos, and so forth) and regress badge outcomes on these labels. Third, what do these patterns teach us about persuasion? By linking calibrated forecasts to strategy coefficients we gain a micro-level view of when evidence overcomes, or fails to overcome, prior commitments.

These questions sit at the intersection of formal epistemology, the philosophy of language, and social epistemology. Methodologically we pair large-scale computational tools with conceptual analysis: the language model functions both as an instrument for tracing statistical regularities in natural argument and as an object whose own predictive commitments invite philosophical scrutiny.

The paper proceeds as follows. Section 2 surveys philosophical accounts of persuasion and the formal limits of Bayesian convergence. Section 3 introduces the Change My View corpus, the forecasting architecture, and our dual strategy-coding pipeline. Section 4 reports empirical findings: baseline model calibration, the added value of strategy features, and illustrative cases. Section 5 interprets those findings, arguing that successful persuasion realises a form of strategically bounded rationality that integrates evidential coherence with credibility cues and minimal empathy. Section 6 concludes.

\section{Theoretical Framework}

Philosophers and theorists have long distinguished between dialectical and rhetorical approaches to persuasion. The dialectical tradition views persuasion as a truth-oriented dialogue: ideally, through reasoned argument and shared evidence, interlocutors should converge on the truth. This perspective has one clear expression in Bayesian frameworks, where belief change is governed by the internal coherence of credences and their updating via Bayesian conditioning. In contrast, the rhetorical tradition emphasizes the art of influence, i.e. how speakers actually persuade audiences, which may involve appeals to emotion, credibility, or style as much as logical argument.

These two traditions sometimes diverge: what ought to persuade a fully rational agent need not always work in practice. Contemporary research has sought to reconcile these traditions. \citet{mercier2011}, in their influential ``argumentative theory of reasoning,'' argue that reasoning evolved not to deliver truth per se, but to persuade others and defend one's views in social contexts. In support of this view, \citet{sperber2010} introduce the concept of epistemic vigilance: a suite of mechanisms evolved to assess the reliability of communicated information and the credibility of speakers.

Yet a tension remains between normative and descriptive accounts of persuasion. Normative theories identify what should rationally persuade: valid arguments, well-supported evidence, and logical coherence. Descriptive theories document what in fact persuades: emotional narratives, credibility displays, and appeals to social identity.

\subsection{Rationality, Agreement and Consensus}

A natural philosophical intuition is that dialogue should lead to common evidence, which in turn should lead to agreement. This expectation is encapsulated in what \citet{nielsen2021} call the optimistic thesis about learning: that rational agents who acquire the same evidence will resolve their disagreements.

This expectation is formalized in theorems like that of \citet{blackwell1962}, which demonstrates that if two Bayesian agents possess non-dogmatic prior beliefs and observe the same complete and infinite sequence of evidence, their posterior beliefs regarding future events will inevitably converge. Meanwhile, \citet{aumann1976}'s agreement theorem establishes that if two rational agents begin with common prior beliefs, and their respective posterior probabilities for a specific event become common knowledge between them, then these posterior probabilities must necessarily be identical. Thus, under sufficiently stringent assumptions, these theorems provide the foundation for a Bayesian account of rational agreement.

\subsection{Higher Order Evidence and Peer Disagreement}

A more flexible approach would be to consider the higher-order evidence available regarding the quality or reliability of one's own or one's peers' evidence and reasoning. There are two camps on how we should incorporate higher order evidence to update our beliefs.

Conciliationists argue that discovering a peer's disagreement should lead one to substantially revise one's own confidence. \citet{feldman2006} contends that in many cases of peer disagreement, the rational response is to suspend judgment or significantly soften one's stance. \citet{christensen2007} similarly argues that knowing a peer disagrees gives one reason to adopt a more humble credence. \citet{elga2007} argues that one should give a disagreeing peer's opinion equal weight to one's own.

On the other hand, \citet{kelly2008} and others put forward the steadfast or Total Evidence View. According to this view, one must account for all the evidence one has, which includes not only the disagreement but also the original reasons that led to one's belief. Kelly argues that if one already has access to the first-order evidence upon which the peer's judgment is based, treating the peer's judgment as additional first-order evidence could constitute ``double-counting''.

Various social epistemological models have attempted to incorporate higher order evidence about other agents. The DeGroot model takes a highly conciliationist flavor: each agent revises her credence by taking a weighted average of her neighbours' current beliefs. The refinement proposed by \citet{lehrer1981} endogenises those weights by letting agents also rate how much others trust them. Models developed by \citet{oconnor2017} and \citet{oconnor2020} incorporate epistemic homophily, by downweighting evidence from agents that disagree with them. This can lead to communities failing to converge and exhibiting long-term polarization.

\subsection{Common Ground}

For persuasion to even get off the ground, the parties must share some common ground. Common ground refers to the set of beliefs, assumptions, and knowledge that speakers mutually recognize as accepted in a conversation \citep{stalnaker2002}. Stalnaker represents common ground by means of a context set: the set of possible worlds in which all propositions belonging to the common ground are true. Communication, particularly through assertion, aims to modify this context set.

\citet{clark2001} liken conversation to a cooperative game whose players must continually coordinate their moves by monitoring---and jointly updating---their ``common ground.'' When a speaker tries to persuade, the aim is not to bulldoze an opponent but to nudge that shared footing. Persuasion, then, can require a skilful reshaping of the coordinates that cooperation requires.

\subsection{Bayesian Multi-belief Models}

Freeborn \citep{freeborn2023,freeborn2024} proposes using a Bayesian model suitable for representing many kinds of higher order evidence. In Freeborn's model, we can represent all of an agent's beliefs with a Bayesian network, with higher up beliefs representing various kinds of background beliefs, including underlying worldviews or ideologies, or higher order beliefs about the reliability of other agents or evidence sources. Such agents may polarize or factionalize, splitting into belief-aligned clusters, despite being fully rational and exposed to identical data.

Freeborn's model assumes significant shared ground between agents. Agents agree on the conditional dependencies in the Bayesian network and on the nature and reliability of the shared evidence. Despite this shared architecture, differences in prior beliefs alone can drive polarization. When evidence is shared, the agents' joint probability distributions grow closer together and the mutual information between them increases. Yet, this very process can drive the marginal probabilities of salient propositions farther apart.

The key insight is that when new evidence arrives, it doesn't speak for itself -- its impact on an agent's belief depends on that agent's prior beliefs, including higher order evidence. Thus the same piece of evidence can push their posterior beliefs in opposite directions. A crucial implication is that adding more evidence does not necessarily resolve disagreement: in fact, it can entrench it. One possible route to resolving such disagreement is deeper alignment on the underlying framework: convergence on higher-order beliefs or background assumptions that govern how evidence is weighed.

\subsection{Computational Approaches to Studying Persuasion}

Computational research on persuasion has typically approached the problem from two angles: the detection and classification of persuasive strategies, and the forecasting of conversational outcomes. Early work by \citet{tan2016} on the ChangeMyView corpus revealed that linguistic alignment and stylistic coordination correlate with successful persuasion.

More recent studies, such as \citet{zeng2024}, extend this line by proposing a comprehensive taxonomy of forty persuasion techniques across fifteen broad strategic categories. Parallel to strategy classification is the emerging field of conversation forecasting. \citet{sicilia2024} demonstrate that pre-trained language models can achieve high forecasting accuracy. They introduce the FortUne Dial benchmark to evaluate model calibration and show that properly tuned small models can outperform much larger counterparts.

These methods rely heavily on the ability of language models to express and manage uncertainty. Work by \citet{kadavath2022}, \citet{lin2022}, and \citet{mielke2022} shows that modern LLMs can represent uncertainty both numerically and linguistically, and that calibration improves predictive reliability.

\subsection{Conceptual Framework for Evaluating Persuasion Success}

Operationalizing ``successful persuasion'' requires translating philosophical and theoretical insights into measurable constructs. In the ChangeMyView corpus, success is indexed by the awarding of a delta badge, a symbolic marker that the original poster has acknowledged a change of view. This signal provides a rare empirical anchor for studying belief change in the wild.

However, not all belief changes are equal. We distinguish between (a) full reversals, where the original poster explicitly abandons their original view; (b) modifications, where they revise or qualify their position; and (c) concessions, where they acknowledge good points without fully changing their mind. To forecast these outcomes, we integrate strategy-based features with LLM-generated probabilities.

\section{Data and Methods}

Our empirical foundation is the ``Winning Arguments'' subset of Reddit's r/ChangeMyView (CMV) community released by \citet{tan2016}. The slice we study contains 3,051 original CMV posts, 293,297 comments and 34,911 speakers. Crucially, each comment carries a delta flag (1 = convinced the OP, 0 = failed, None = context) providing an operational marker of successful persuasion at scale.

Note that CMV users self-select for good-faith debate and English fluency; posts on polarising political topics or non-Western issues are under-represented. Moreover, the delta is awarded by the OP alone, an idiosyncratic, binary signal that almost certainly understates smaller doxastic shifts. We therefore treat all findings as conservative with respect to persuasion frequency and emphasise calibration metrics rather than raw accuracy.

To emulate the real-time uncertainty faced by discussants, we follow \citet{sicilia2024}'s ``partial-conversation'' protocol: for every target thread we truncate the dialogue at a random intermediate turn $K$ and ask a forecaster whether a delta will eventually be awarded.

Using LLMs both to read arguments and to forecast their success challenges a sharp division between descriptive statistics and first-order participation. On one hand the baseline predictor functions as a Bayesian-like ideal observer, aggregating lexical, pragmatic and social cues into well-calibrated credences. On the other, the strategy pipeline reveals that these credences track recognisable rhetorical moves.

\subsection{Forecasting Method 1: Baseline LLM}

We adapt the Theory-of-MindGPT prompt of \citet{sicilia2024}. A system message casts the model as ``an expert at predicting the beliefs and actions of others''; the user message embeds (i) the OP's title and body, (ii) the unfolding reply, and (iii) the question ``Will the original poster change their opinion? Let's think step-by-step \ldots Answer on a 1 -- 10 scale.'' The model's numeric answer is parsed via regex and linearly mapped to a probability $p = (k - 1)/9$.

We experiment with Llama-3 8B and 70B in their instruction-tuned variants. Predictions are evaluated with:
\begin{itemize}
\item Brier score (mean-squared error between $p$ and outcome $y$),
\item Brier skill score (improvement over the climatological baseline),
\item F1 (after thresholding at 0.5),
\end{itemize}

A temperature of 0.7 and top-p = 0.9 were used for generation. The baseline forecasts were also evaluated with output scaling, using a small amount of data to adjust the model outputs. This procedure is distinct from the fitting of the strategy-based logistic regression models; the baseline prompting procedure does not update the language model weights.

\subsection{Forecasting Method 2: Strategy-Aware Models}

Persuasion tactics are labelled by an LLM-assisted pipeline:
\begin{enumerate}
\item \textbf{Free elicitation.} For each response we prompt StrategyClassifierGPT to list verb-phrase strategies ``in one or two words''. 
\item \textbf{Taxonomy induction.} The resulting 1,700+ raw phrases are grouped by a further model pass into super-categories. If phrases remain unassigned, the model is prompted again with the missing phrases. Authors then inspect and refine the resulting categories, including combining, reorganizing, and removing sparsely populated categories (fewer than ten instances); a final model pass provides descriptions.
\item \textbf{Fixed-taxonomy coding.} Each response is classified again using the resulting ten defined super-categories, generating the binary strategy indicators used in the regressions.
\end{enumerate}

We estimate three progressively richer logistic models that relate strategy use to persuasion success. Let $\mathbf{s}$ denote the 10-element binary vector marking whether each super-strategy is present in the target reply, and let $p$ be the baseline LLM probability:
\begin{itemize}
\item \textbf{Strategy-Only model (S-only):} Tests whether strategy cues alone carry predictive power.
\item \textbf{Additive model (LLM + S):} Here the raw LLM forecast is treated as another covariate.
\item \textbf{Interaction model (LLM $\times$ S):} The element-wise product $p\mathbf{s}$ lets us measure how much a given strategy conditions the credibility of the LLM's prior odds.
\end{itemize}

All coefficients are estimated by maximum likelihood with conversation-level 10-fold cross-validation. We report mean Brier score and $\Delta$Brier for ablation of each term. Two training-set sizes are considered: standard ($n = 800$) and big-data ($n = 4,000$).

\section{Results}

\subsection{Baseline Forecasting Performance}

Our baseline model employs a chain-of-thought prompting strategy in which a language model predicts whether a delta will eventually be awarded. We compare unscaled forecasts with forecasts after output scaling using a small amount of data. For Llama 3.1 70B, scaling reduces the Brier score from 0.312 to 0.238 and increases F1 from 0.555 to 0.776. For Llama 3.1 8B, it reduces the Brier score from 0.282 to 0.236 and increases F1 from 0.637 to 0.765. These changes concern scaled forecasts, not language-model weight fine-tuning.

The corresponding Brier skill scores, measured against the reference forecast used in the evaluation, improve from $-0.350$ to $-0.030$ for 70B and from $-0.222$ to $-0.022$ for 8B. Thus scaling improves the reported scores, but the negative skill scores indicate that both scaled forecasts remain worse by this metric than the reference forecast. We do not infer calibration quality from these aggregate scores alone.

\subsection{Strategy Analysis Results}

To examine the content of persuasion, we annotated CMV replies with argumentative strategies. Two taxonomies were used: one derived from social influence literature (the ``Lit'' taxonomy) and one induced via chain-of-thought prompting with clustering and author curation (the ``LLM'' taxonomy).

Both taxonomies converge on core rhetorical categories, but the LLM-induced taxonomy demonstrates tighter clustering and slightly better predictive power. The complete set of identified categories includes:

\begin{itemize}
\item \textbf{Establishing credibility and authority:} Citing evidence, referencing authority, quoting sources, establishing expertise.
\item \textbf{Challenging assumptions and arguments:} Pointing out flaws, dismissing arguments, questioning authority.
\item \textbf{Building empathy and rapport:} Showing empathy, acknowledging emotions, expressing gratitude, humanizing.
\item \textbf{Providing alternative perspectives:} Offering explanations, counterexamples, reframing issues.
\item \textbf{Using logical reasoning and evidence:} Using analogies, data, statistics, appealing to logic.
\item \textbf{Emotional appeals and manipulation:} Using emotional language, personal attacks, rhetorical questions.
\item \textbf{Deflecting and diverting attention:} Deflecting criticism, changing the topic, dismissing views.
\item \textbf{Setting boundaries and limitations:} Setting limits, establishing exceptions, narrowing options.
\item \textbf{Reframing and redefining:} Redefining terms, recontextualizing, changing issue framing.
\item \textbf{Conceding and compromising:} Conceding points, offering compromises, finding common ground.
\end{itemize}

Distributional analysis reveals that logical reasoning is the most common strategy, followed by empathy-building and concession. Some categories (e.g., emotional appeals and deflection) were rare but still occasionally predictive of success.

\subsection{Predictive Power of Different Strategies}

To evaluate how specific strategies predict persuasion, we trained three logistic regression models. In the strategy-only model, several strategies showed statistically significant positive coefficients for persuasion success. The interaction model consistently outperformed the additive model, suggesting that strategy efficacy is not merely additive but conditional. The results in the most predictive model are displayed in Table~\ref{tab:results}.

\begin{table}[h]
\centering
\caption{Coefficients and p-values for the logistic regression model with the LLM-induced taxonomy and interaction effects included.}
\label{tab:results}
\begin{tabular}{lcc}
\toprule
\textbf{Strategy} & \textbf{Coefficient} & \textbf{p-value} \\
\midrule
\multicolumn{3}{l}{\textit{Positively Significant (p < 0.05)}} \\
Conceding and compromising & +0.86 & < 0.001 \\
Building empathy and rapport & +0.39 & < 0.001 \\
\midrule
\multicolumn{3}{l}{\textit{Insignificant (p $\geq$ 0.05)}} \\
Establishing credibility and authority & +0.07 & 0.298 \\
Emotional appeals and manipulation & +0.02 & 0.805 \\
Setting boundaries and limitations & -0.10 & 0.519 \\
Reframing and redefining & -0.11 & 0.086 \\
Providing alternative perspectives & -0.14 & 0.157 \\
\midrule
\multicolumn{3}{l}{\textit{Negatively Significant (p < 0.05)}} \\
Using logical reasoning and evidence & -0.16 & 0.029 \\
Challenging assumptions \& arguments & -0.37 & < 0.001 \\
Deflecting and diverting attention & -0.48 & < 0.001 \\
\bottomrule
\end{tabular}
\end{table}

\subsection{Interpretation and Limitations}

The strongest predictors of persuasion success combine strategic structure with appropriately modulated confidence. Strategy-aware interaction models outperform both naïve baselines and simple additive models, suggesting that successful argumentation involves rhetorical synergy.

Despite the strength of our findings, several limitations deserve note. First, our strategy annotations were generated primarily by large language models, with limited human oversight. While this approach enabled broad coverage and consistency, it introduces potential inaccuracies. The category labeled ``logical reasoning and evidence,'' for example, may capture instances of mere mention rather than effective use.

Second, our operational definition of persuasion---whether a delta badge was awarded---is necessarily coarse. It captures only explicit, acknowledged belief change and likely underrepresents partial, private, or delayed shifts.

Third, the use of LLMs in both predicting outcomes and identifying strategies raises concerns about model circularity. Although our design includes safeguards, there remains a risk that the model's own priors shape both features and labels.

Fourth, the Reddit CMV corpus is subject to selection biases. Reddit's demographic profile is not fully representative, and CMV users specifically self-select for openness to persuasion and fluency in English and digital discourse norms.

Finally, the structured nature of ChangeMyView may limit generalizability. The platform's strong moderation and emphasis on good-faith dialogue foster unusually cooperative exchanges.

Despite these caveats, our results offer a demonstration of how computational tools can model persuasion dynamics at scale and begin to illuminate the rhetorical scaffolding that underlies belief change.

\section{Philosophical Discussion}

The regression coefficients reported in Section 4 may appear concerning at first glance, especially for advocates of rational persuasion. In the conversational ecology of r/ChangeMyView, replies that begin with concessionary language or that make a visible effort to establish rapport are markedly more likely to earn a delta than replies that lead with formal refutation or a battery of facts.

The asymmetry aligns neatly with the two traditions reviewed in Section 2. The dialectical ideal predicts that well-supported evidence should eventually persuade any rational agent with compatible priors. The rhetorical tradition emphasises ethos and pathos---credibility displays, empathy, identity cues---as indispensable levers of influence.

Our data endorse the rhetorical insight without rejecting the dialectical one. Empathy-building and concession can be understood as socially-adept precursors to explicit reasoning, fulfilling the pragmatic conditions under which evidence can be admitted and processed. This echoes \citet{mercier2011}'s ``argumentative theory of reasoning,'' according to which the primary evolutionary function of reason is to persuade others within a cooperative group. Epistemic vigilance mechanisms \citep{sperber2010} make audiences wary of bare assertions; relational cues help the speaker pass those filters.

The data also speak to the debate between conciliationist and steadfast responses to peer disagreement. Our findings suggest a middle path: concessionary dialogue may facilitate rational disagreement resolution by helping the parties assess evidence under more closely aligned interpretive assumptions. Temporary softening (a willingness to concede small points) is instrumental to a more reliable eventual verdict on the first-order question.

However, this pattern should not be misread as a victory of sentiment over reason. The speaker must first lower interpersonal barriers: acknowledging the interlocutor's concerns, signalling goodwill, perhaps offering a modest concession. Recall that \citet{stalnaker2002} models conversation in terms of a shared context set. Establishing common ground may help the hearer incorporate the proposed content into that context.

Our results clarify a mechanism by which that incorporation may happen. Concession is a public demonstration that speaker and hearer already share part of the context set; it literally enlarges the overlap of accepted propositions. Empathy performs a similar function at the level of attitudes and values. These moves reduce the epistemic risk that the hearer faces when contemplating belief revision.

How, then, should we interpret the negative coefficients for logical challenge and evidential bombardment? The Bayesian-network model developed by Freeborn \citep{freeborn2023,freeborn2024} offers a ready explanation. In his Bayesian-network formalism, each agent's belief state is a web of probabilistic dependencies. When a new datum arrives, its impact on some target proposition $P$ is mediated by intermediate nodes---background beliefs, reliability attributions, identity markers. Two agents can therefore update rationally on the same evidence and nonetheless move farther apart on $P$ if their inter-belief wiring differs.

A head-on logical challenge may engage precisely those mediating beliefs on which agents diverge. Under suitable conditions, the resulting Bayesian updates can reinforce rather than reduce disagreement. By contrast, concessionary or empathetic openings may work indirectly to align portions of the network higher up---pruning scepticism about the speaker's intentions, signalling shared values, or reframing source reliability. Once that alignment occurs, subsequent factual claims have a greater chance of being routed through comparable internal pathways, yielding convergence rather than divergence.

Our CMV results are consistent with this picture: comments that begin by attacking an opponent's reasoning may inadvertently activate precisely those background pathways where priors diverge, leading to entrenchment. Conversely, rapport-oriented moves might dampen the weight of those divergent pathways, thereby making later evidence more likely to be interpreted through overlapping segments of the network.

Seen in this light, emotional or common-ground strategies need not be viewed as irrational substitutes for reason. Instead, they can be interpreted as conversational tactics that modulate the higher-order channels through which evidence flows.

\section{Conclusions}

Our study contributes a scalable, computationally grounded method for investigating the micro-dynamics of belief change in online dialogue. Our findings open several promising directions for refining models of persuasion and extending their philosophical and practical significance.

Future work should capture more subtle forms of belief change beyond binary delta awards, such as softening, hedging, or partial concession, through richer linguistic markers. Our taxonomy could also be expanded: disaggregating broad categories may reveal topic-sensitive sub-strategies. Modeling topic effects more explicitly may help clarify when and why certain strategies succeed.

Our results invite reconsideration of rational belief change under socially situated conditions. Strategies like concession and credibility displays, while pragmatically effective, may also be epistemically virtuous by fostering uptake and trust. More broadly, our findings reinforce the view that polarization can emerge from rational inference over entangled priors, rather than epistemic failure.

Insights from our models could inform educational tools for teaching effective argumentation, as well as platform designs that promote productive discourse. However, any tool that forecasts persuasion raises ethical concerns: predictive models could be weaponized for manipulation. Future work must develop safeguards to align persuasive technologies with democratic and epistemic norms.

Finally, our micro-level results can be embedded into simulations of networked belief dynamics. Agent-based models could test whether strategically ``seeding'' conversational styles affects overall polarization. This would link individual-level discourse features with broader phenomena such as echo chambers, misinformation spread, or epistemic resilience.

\bibliographystyle{apalike}
\bibliography{references}

@article{aumann1976,
  title={Agreeing to Disagree},
  author={Aumann, Robert J},
  journal={The Annals of Statistics},
  volume={4},
  number={6},
  pages={1236--1239},
  year={1976}
}

@article{blackwell1962,
  title={Merging of Opinions with Increasing Information},
  author={Blackwell, David and Dubins, Lester},
  journal={The Annals of Mathematical Statistics},
  volume={33},
  number={3},
  pages={882--886},
  year={1962}
}

@incollection{clark2001,
  title={Grounding in Communication},
  author={Clark, Herbert H and Brennan, Susan E},
  booktitle={Perspectives on Socially Shared Cognition},
  editor={Resnick, Lauren B and Levine, John M and Teasley, Stephanie D},
  pages={127--149},
  year={1991},
  publisher={American Psychological Association}
}

@article{christensen2007,
  title={Epistemology of Disagreement: The Good News},
  author={Christensen, David},
  journal={The Philosophical Review},
  volume={116},
  number={2},
  pages={187--217},
  year={2007}
}

@article{elga2007,
  title={Reflection and Disagreement},
  author={Elga, Adam},
  journal={No\^{u}s},
  volume={41},
  number={3},
  pages={478--502},
  year={2007}
}

@phdthesis{freeborn2023,
  title={Polarization and Factionalization for Agents with Multiple, Related Beliefs},
  author={Freeborn, David Peter Wallis},
  year={2023},
  school={University of California, Irvine}
}

@article{freeborn2024,
  title={Rational Factionalization for Agents with Probabilistically Related Beliefs},
  author={Freeborn, David Peter Wallis},
  journal={Synthese},
  volume={203},
  number={2},
  pages={1--27},
  year={2024}
}

@article{kelly2008,
  title={Disagreement, Dogmatism, and Belief Polarization},
  author={Kelly, Thomas},
  journal={The Journal of Philosophy},
  volume={105},
  number={10},
  pages={611--633},
  year={2008}
}

@article{mercier2011,
  title={Why Do Humans Reason? {A}rguments for an Argumentative Theory},
  author={Mercier, Hugo and Sperber, Dan},
  journal={Behavioral and Brain Sciences},
  volume={34},
  number={2},
  pages={57--74},
  year={2011}
}

@misc{marist2018sept,
  title={{NPR/PBS NewsHour/Marist Poll National Tables, September 2018}},
  author={{Marist Poll}},
  year={2018},
  howpublished={Marist College Institute for Public Opinion}
}

@misc{marist2018oct,
  title={{NPR/PBS NewsHour/Marist Poll National Tables, October 2018}},
  author={{Marist Poll}},
  year={2018},
  howpublished={Marist College Institute for Public Opinion}
}

@inproceedings{sicilia2024,
  title={Deal, or No Deal (or Who Knows)? {F}orecasting Uncertainty in Conversations Using Large Language Models},
  author={Sicilia, Anthony and Kim, Hyunwoo and Chandu, Khyathi Raghavi and Alikhani, Malihe and Hessel, Jack},
  booktitle={Findings of the Association for Computational Linguistics: ACL 2024},
  pages={11700--11726},
  year={2024},
  publisher={Association for Computational Linguistics}
}

@article{sperber2010,
  title={Epistemic Vigilance},
  author={Sperber, Dan and Cl\'{e}ment, Fabrice and Heintz, Christophe and Mascaro, Olivier and Mercier, Hugo and Origgi, Gloria and Wilson, Deirdre},
  journal={Mind \& Language},
  volume={25},
  number={4},
  pages={359--393},
  year={2010}
}

@article{stalnaker2002,
  title={Common Ground},
  author={Stalnaker, Robert},
  journal={Linguistics and Philosophy},
  volume={25},
  number={5--6},
  pages={701--721},
  year={2002}
}

@inproceedings{tan2016,
  title={Winning Arguments: Interaction Dynamics and Persuasion Strategies in Good-Faith Online Discussions},
  author={Tan, Chenhao and Niculae, Vlad and Danescu-Niculescu-Mizil, Cristian and Lee, Lillian},
  booktitle={Proceedings of the 25th International Conference on World Wide Web},
  pages={613--624},
  year={2016}
}

@article{nielsen2021,
  author = {Nielsen, Michael and Stewart, Rush T.},
  title = {Persistent Disagreement and Polarization in a Bayesian Setting},
  journal = {The British Journal for the Philosophy of Science},
  volume = {72},
  number = {1},
  pages = {51--78},
  year = {2021},
  doi = {10.1093/bjps/axy056}
}

@incollection{feldman2006,
  author = {Feldman, Richard},
  title = {Epistemological Puzzles about Disagreement},
  booktitle = {Epistemology Futures},
  year = {2006},
  pages = {199--215},
  publisher = {Oxford University Press},
  doi = {10.1093/oso/9780199273317.003.0013}
}

@book{lehrer1981,
  author = {Lehrer, Keith and Wagner, Carl},
  title = {Rational Consensus in Science and Society: A Philosophical and Mathematical Study},
  year = {1981},
  publisher = {D. Reidel},
  address = {Dordrecht}
}

@article{oconnor2017,
  author = {O'Connor, Cailin and Weatherall, James Owen},
  title = {Scientific Polarization},
  journal = {European Journal for Philosophy of Science},
  volume = {8},
  pages = {855--875},
  year = {2018},
  doi = {10.1007/s13194-018-0213-9},
  note = {First circulated as a preprint in 2017}
}

@article{oconnor2020,
  author = {Weatherall, James Owen and O'Connor, Cailin},
  title = {Endogenous Epistemic Factionalization},
  journal = {Synthese},
  year = {2021},
  volume = {198},
  pages = {6179--6200},
  doi = {10.1007/s11229-020-02675-3}
}

@inproceedings{zeng2024,
  author = {Zeng, Yi and Lin, Hongpeng and Zhang, Jingwen and Yang, Diyi and Jia, Ruoxi and Shi, Weiyan},
  title = {How Johnny Can Persuade {LLMs} to Jailbreak Them: Rethinking Persuasion to Challenge {AI} Safety by Humanizing {LLMs}},
  booktitle = {Proceedings of the 62nd Annual Meeting of the Association for Computational Linguistics (Volume 1: Long Papers)},
  year = {2024},
  pages = {14322--14350},
  doi = {10.18653/v1/2024.acl-long.773},
  url = {https://aclanthology.org/2024.acl-long.773/}
}

@misc{kadavath2022,
  author = {Kadavath, Saurav and others},
  title = {Language Models (Mostly) Know What They Know},
  year = {2022},
  eprint = {2207.05221},
  archivePrefix = {arXiv},
  url = {https://arxiv.org/abs/2207.05221}
}

@article{lin2022,
  author = {Lin, Stephanie and Hilton, Jacob and Evans, Owain},
  title = {Teaching Models to Express Their Uncertainty in Words},
  journal = {Transactions on Machine Learning Research},
  year = {2022},
  url = {https://mlanthology.org/tmlr/2022/lin2022tmlr-teaching/}
}

@article{mielke2022,
  author = {Mielke, Sabrina J. and Szlam, Arthur and Dinan, Emily and Boureau, Y-Lan},
  title = {Reducing Conversational Agents' Overconfidence Through Linguistic Calibration},
  journal = {Transactions of the Association for Computational Linguistics},
  volume = {10},
  pages = {857--872},
  year = {2022},
  doi = {10.1162/tacl_a_00494}
}

\end{document}